\documentclass[10pt,twocolumn,letterpaper]{article}

\usepackage{cvpr}
\usepackage{times}
\usepackage{epsfig}
\usepackage{graphicx}
\usepackage{amsmath}
\usepackage{amssymb}

\usepackage{amsmath}
\usepackage{graphics}
\usepackage{graphicx}
\usepackage{amssymb}
\usepackage{algorithm}
\usepackage{algorithmic}
\usepackage{mathrsfs}
\usepackage{booktabs}
\usepackage{bm}
\usepackage{bbm}
\usepackage{float}
\usepackage{url}
\usepackage{amsfonts}
\usepackage[square,sort&compress,numbers]{natbib}
\usepackage{multirow}
\usepackage{textcomp}
\usepackage{supertabular}
\usepackage{array}
\usepackage{caption}
\usepackage{gensymb}
\usepackage{threeparttable}
\usepackage{makecell}
\usepackage{amsmath}
\usepackage{setspace}

\usepackage[breaklinks=true,bookmarks=false]{hyperref}

\cvprfinalcopy 

\def\cvprPaperID{} 

\ifcvprfinal\fi
\begin{document}

\urlstyle{rm}

\newcolumntype{L}[1]{>{\raggedright\arraybackslash}p{#1}}
\newcolumntype{C}[1]{>{\centering\arraybackslash}p{#1}}
\newcolumntype{R}[1]{>{\raggedleft\arraybackslash}p{#1}}

\title{ADCPNet: Adaptive Disparity Candidates Prediction Network for \\Efficient Real-Time Stereo  Matching}

\author{
He Dai,\ \ Xuchong Zhang,\ \ Yongli Zhao,\ \ Hongbin Sun\\
Xi'an Jiaotong University
}

\maketitle

\begin{abstract}
	Efficient real-time disparity estimation is critical for the application of stereo vision systems in various areas. Recently,  stereo network based on coarse-to-fine method has largely relieved the memory constraints and speed limitations of large-scale network models.  Nevertheless, all of the previous coarse-to-fine designs employ constant offsets and three or more stages to progressively refine the coarse disparity map, still resulting in unsatisfactory computation accuracy and inference time when deployed on mobile devices. This paper claims that the coarse matching errors can be corrected efficiently with fewer stages as long as more accurate disparity candidates can be provided. Therefore, we propose a dynamic offset prediction module to meet different correction requirements of diverse objects and design an efficient two-stage framework. Besides, we propose a disparity-independent convolution to further improve the performance since it is more consistent with the local statistical characteristics of the compact cost volume. The evaluation results on multiple datasets and platforms clearly demonstrate that, the proposed network outperforms the state-of-the-art lightweight models especially for mobile devices in terms of accuracy and speed. Code will be made available.
\end{abstract}


\section{Introduction}
\begin{figure}[t]
	\centering
	\includegraphics[width=0.9\columnwidth]{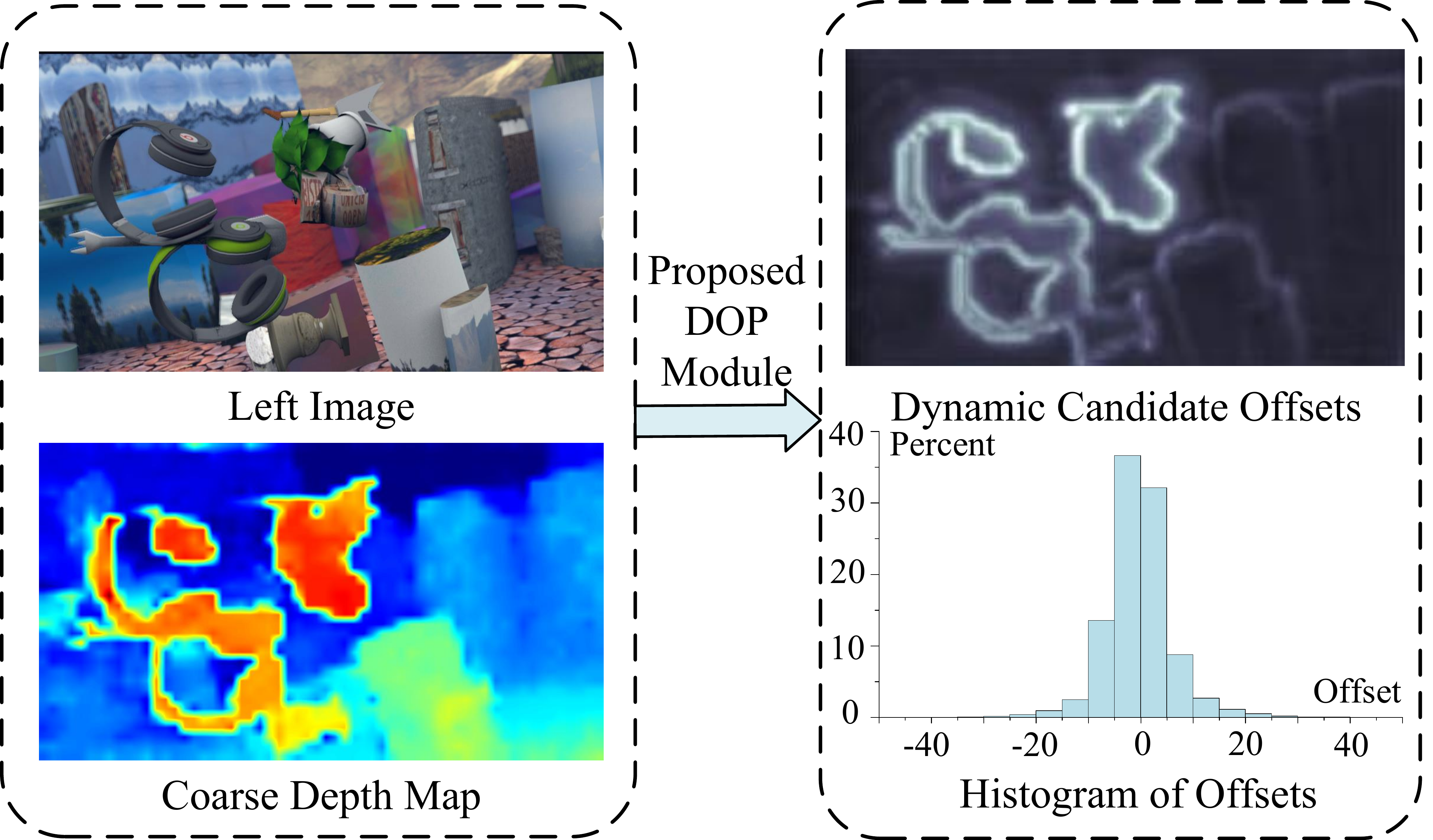} 
	\caption {The visualization results of our predicted offsets and histogram statistics of one sample. The offset range is much wider than that of traditional constant offset ($\pm$1, $\pm$2) in order to accommodate different requirements of disparity correction for nearby, distant, thin or small objects. }
	\label{fig1}
\end{figure}
Stereo vision system has been widely used in a wide variety of application areas, including robot navigation, intelligent surveillance and autonomous driving~\cite{luo2017vision, chen20153d,sivaraman2013review}. Therefore, efficient and accurate disparity estimation on mobile devices is practically interesting and crucial to its real-time deployment. 

Owing to the great success of deep convolutional neural network (CNN),  learning-based stereo networks~\cite{kendall2017end,yu2018deep,mayer2016large,vzbontar2016stereo,liang2019unsupervised} have enabled remarkable progress compared to conventional local or global methods~\cite{besse2013pmbp,scharstein2001taxonomy} in terms of occlusion areas, repeated patterns, large textureless regions and reflective surfaces.
Following the traditional stereo matching pipeline, an end-to-end stereo network usually contains feature learning, cost volume construction, cost aggregation and disparity regression / refinement. 
A great majority of previous works focus on designing elaborate network models in these steps to continuously improve disparity accuracy. 
However, the state-of-the-art (SOTA) networks usually require a large amount of memory and computation resources to obtain accurate disparity results, which makes them difficult to deploy on source-constrained platforms. As presented in~\cite{zhang2019ga}, GANet takes 6.2GB memory and 1.8s to inference KITTI image pairs even on high-end GPU. 

The cost aggregation step largely determines the performance and efficiency of the overall network. We note that the current SOTA works generally exploit 3D CNN to regularize the concatenated cost volume. Nevertheless, the 3D convolution is inherently expensive since the cost volume is a 4D tensor which consists of \textit{height} $\times$ \textit{width} $\times$ \textit{disparity range} $\times$ \textit{feature size}. Recently, a few studies have been proposed to reduce the computation complexity of large-scale stereo networks by employing the following techniques. (1) Further downsampling the spatial resolution of the feature map to construct a small cost volume~\cite{khamis2018stereo}. (2) Compressing the feature size of cost volume by matching module~\cite{tulyakov2018practical} or correlation layer~\cite{mayer2016large}.  (3) Exploiting coarse-to-fine method to avoid full-range disparity computation~\cite{yan2019anytime,dovesi2019real,tonioni2019real}. Compared with the former two techniques, the coarse-to-fine method provides a more efficient and lightweight solution to relieve the memory constraints
or speed limitations especially for mobile devices (e.g. NVIDIA Jetson TX2). However, we observe that  all of the previous coarse-to-fine stereo networks correct the matching errors in high resolution by adding a few constant offsets to the  coarse disparity results regressed in low resolution. Since the constant values limit the disparity candidates in current stage only to a local range around the coarse results of the previous stage, it is difficult to meet the practical requirements of various objects in different scenes. As it can be seen, objects that appear at different depths are projected to different scales in the disparity map, thus the disparity errors of nearby and distant objects actually need to be fine tuned with different offset ranges rather than constant ones.
In addition, the disparity corrections of small or thin objects may need totally different candidates with the coarse results due to the information loss incurred by the downsampling and upsampling operations. 
Therefore, the computation accuracy of previous coarse-to-fine models for mobile applications still has a large gap compared to large-scale SOTA stereo networks. Moreover, to improve the computation accuracy, the previous designs usually exploit three or more stages to progressively refine the coarse disparity map, but the multi-stage pyramid processing decreases the computation speed significantly as the number of stages increases. Hence, the real-time performance of previous stereo networks is still unsatisfactory especially for mobile devices.

In this paper, we claim that the coarse matching errors can be corrected efficiently with fewer stages as long as  more accurate disparity candidates can be provided in the fine disparity estimation stage. 
Accordingly, we firstly propose a dynamic offsets prediction (DOP) module to generate adaptive disparity candidates
in order to accommodate different requirements of disparity correction for various objects (Figure \ref{fig1}). In addition, a disparity-independent convolution (DIC) is proposed to further improve the overall accuracy and speed since it presents a more efficient regularization for the compact cost volume in the fine disparity estimation stage. Based on these two simple yet efficient modules, we design a two-stage coarse-to-fine architecture, namely Adaptive Disparity Candidates Prediction Network (ADCPNet) to achieve a better balance between computation accuracy and inference time for efficient real-time stereo matching. 
The real-time performance of the proposed ADCPNet is extensively evaluated on multiple datasets and GPU platforms. Compared with the previous coarse-to-fine designs for mobile devices, the proposed network achieves better disparity accuracy even without extra refinement operation.  Moreover, our design achieves higher efficiency for the final disparity estimation and doubles the computation speed of the current SOTA lightweight stereo networks either on source-constrained or high-end GPU in the case of similar disparity accuracy. The evaluation results clearly demonstrate the effectiveness and efficiency of the proposed stereo network. 
\section{Related Work}

\subsection{Deep learning based stereo matching}
Deep learning based stereo matching algorithms achieve
significant accuracy improvement compared with hand-crafted methods. CNN is initially embedded into conventional stereo matching flow to systematically learn good features from image patches~\cite{vzbontar2016stereo,zhang2017fundamental,seki2017sgm}. These methods improve the traditional disparity accuracy, but are still challenged in ill-posed regions.
DispNetC~\cite{mayer2016large} is the first end-to-end trainable stereo network, which employs a correlation layer to construct cost volume and regularize it by a stack of 2D convolutions. GCNet~\cite{kendall2017end} largely boosts the performance by preserving the feature dimension to release the ability of 3D CNN on a concatenate-based 4D cost volume. 
Based on GCNet, PSMNet~\cite{chang2018pyramid} takes the advantage of spatial pyramid architecture to extract multi-scale contextual information. 
AcfNet~\cite{you2020adaptive} supervises the cost volume with ground truth unimodal distributions to solve the under-constrain overfitting problem.
Recently, GANet~\cite{zhang2019ga} and CSPN~\cite{cheng2019learning} apply the classical long-range spatial propagation scheme in feature learning and cost aggregation steps to obtain superior accuracy. 

SOTA stereo networks obtain high quality disparity map at the cost of a large number of parameters, FLOPs and memory overhead, making them impractical to be integrated into mobile applications, such as smart phones or self-driving cars. Therefore, a growing trend is to design lightweight and
efficient architecture with reasonable disparity estimation for stereo network.



\subsection{Real-time stereo networks}
\begin{figure*}[!t]
	\centering
	\includegraphics[width=1.9\columnwidth]{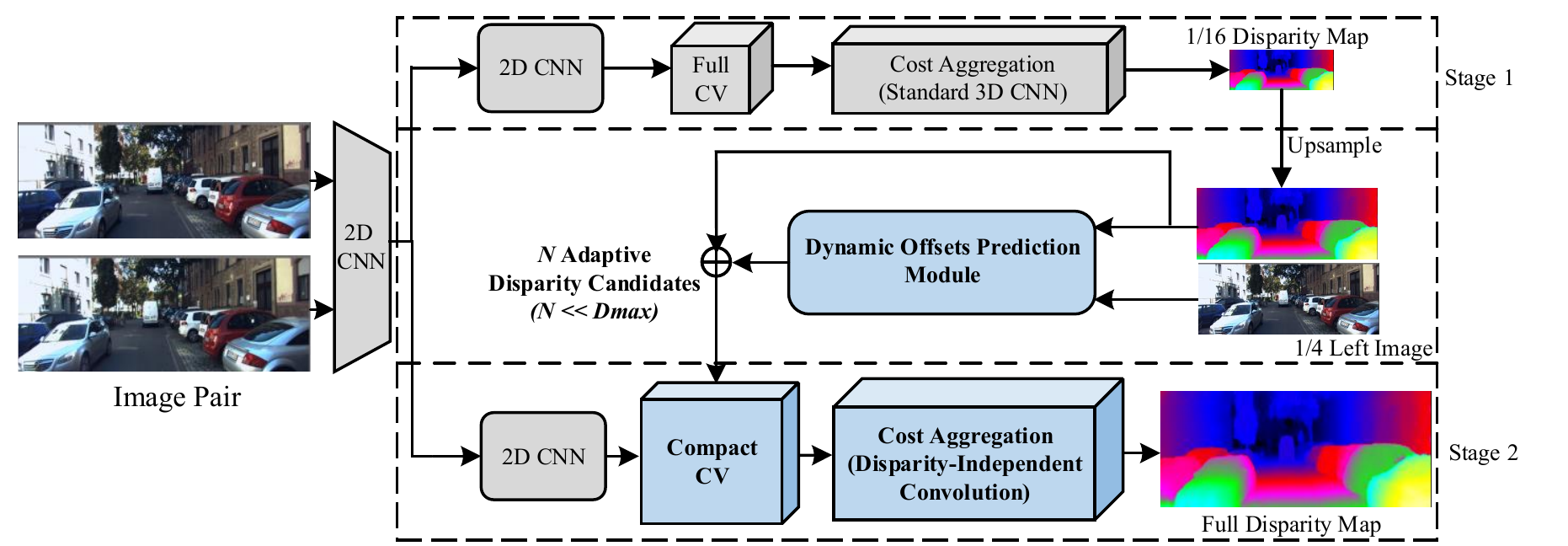} 
	\caption{Overall architecture of the proposed Adaptive Disparity Candidates Prediction Network (ADCPNet). We design a two-stage coarse-to-fine framework and propose two efficient modules to improve the accuracy and speed, where the dynamic offset prediction is used to predict a few offsets dynamically and then generate the adaptive disparity candidates for each pixel, and the  disparity-independent convolution is used to regularize the compact cost volume more efficiently.}
	\label{fig4}
\end{figure*}
As introduced in the above Section, the expensive 3D CNN used to regularize the 4D cost volume significantly increases the computation complexity of SOTA stereo network. StereoNet~\cite{khamis2018stereo} is the first end-to-end architecture for real-time implementation which aggregates the cost volume in a low-resolution with a downsampling factor of 8. Although it runs at 60 FPS on Titan X, the disparity accuracy is unsatisfactory even refined by an edge-aware upsampling function. PDS~\cite{tulyakov2018practical} proposes  a bottleneck matching module to compress the feature size of cost volume. Since this method keeps the dimension of resolution and disparity search range, the computation speed is relatively slower than other designs. DeepPruner~\cite{duggal2019deep} prunes out the search space by exploiting a differentiable PatchMatch algorithm and constructing a sparse representation of the cost volume, resulting in a 16 FPS frame rate on Titan Xp.

AnyNet~\cite{yan2019anytime}, RTS$^2$~\cite{dovesi2019real} and  MADNet~\cite{tonioni2019real} are the existing three networks aiming at the real-time deployment on the low power TX2 platform. All of these designs employ a  multi-stage ($\ge3 $ stages) coarse-to-fine strategy to reduce the computation complexity and memory footprint. Specifically, these architectures firstly regress a coarse disparity map by regularizing the full-range cost volume in low resolution. Then, the coarse disparity map is progressively upsampled and used to warp the features at high resolution.  Since the coarse disparities contain a lot of mismatches especially in discontinuous regions, five constant offsets ($-2,-1,0,1,2$) are utilized to generate more disparity candidates for the result correction in the fine disparity estimation stages.

This paper also employs the coarse-to-fine method to satisfy the real-time performance of mobile applications. However, our work  differs from the above models in the following three aspects. (1) We propose an adaptive prediction module instead of the existing constant offsets to adapt different correction ranges in different image positions. (2) We propose a disparity-independent convolution (i.e. weight-unshared CNN)  rather than using the traditional 3D CNN to regularize the compact cost volume. (3) Because of the effectiveness of the proposed two modules, we employ a two-stage coarse-to-fine design which enables both the accuracy and speed.
In the following section,  we detail our overall network architecture and each critical component.

\section{ADCPNet}
\subsection{Network overview}
The overall architecture of the proposed ADCPNet is shown in Figure \ref{fig4}. Given a rectified image pair, we firstly extract
downsampled feature maps at 1/2, 1/4, 1/8 and 1/16 resolutions using a series of residual blocks~\cite{he2016deep} with stride of 2, where the feature size of each scale is set as $2C$, $2C$, $4C$ and $8C$ respectively.  $C$ is a constant value which determines the performance and computation complexity of feature extractor.
Then, a two-stage coarse-to-fine model is exploited to achieve efficient and accurate disparity maps.

In stage 1, two residual blocks with stride of 1 are cascaded to learn the unary feature in 1/16 resolution. A full-range cost volume is constructed by concatenating the left image
feature with its corresponding right image feature across disparity search range (i.e. 0 $\sim$ $D_{max}/16$). Afterwards, we use six standard 3D convolutions with the same feature size $C_{3d}$ to regularize the cost volume. The coarse disparity map is finally obtained by  the soft argmin function and regression scheme~\cite{kendall2017end}. It should be noted that we add an intermediate disparity output after the first layer of 3D convolutions to provide an extra supervision during the training phase.

The processing flow in stage 2 is similar with stage 1.  The major differences between these two stages are the disparity candidates generation that is used to construct cost volume and the convolution operations that are used for cost aggregation. Specifically,  we propose a dynamic offsets prediction scheme and a disparity-independent convolution for the fine disparity regression. The  detailed structure of the proposed ADCPNet is listed in supplementary materials, and the following subsections describe the proposed two modules and the loss function individually.
\subsection{Dynamic offsets prediction}
After obtaining the results in stage 1, the coarse-to-fine strategy constructs a compact cost volume in the following stage by appending a set of offsets to the coarse disparity for each pixel. Formally, the disparity candidates for fine disparity estimation can be defined as
\begin{equation}
	\begin{aligned}
		\mathbf{dc_{p}}=\{d_{\mathbf{p}} + k_{\mathbf{p}}^n |n=1, 2,  ..., N\}.
	\end{aligned}
	\label{e0}
\end{equation}
Where $\mathbf{dc_{\mathbf{p}}}$ is the candidate set for pixel $\mathbf{p}$. $d_{\mathbf{p}}$ and $k_{\mathbf{p}}^n$ indicate the coarse disparity of pixel $\mathbf{p}$ and its $n$th offset respectively. $N$ is the number of offsets. Since each pixel only holds a few disparity candidates instead of all possible ones in high resolution, i.e., $N \ll D_{max}$, the  coarse-to-fine method enables substantial complexity reduction and computation speedup.

In all previous works~\cite{yan2019anytime,dovesi2019real,tonioni2019real},  the offsets are set as constant values, which can be formulated as
\begin{equation}
	\begin{aligned}
		k_{\mathbf{p}}^n=n - \lceil {N \over  2} \rceil.
	\end{aligned}
	\label{e1}
\end{equation}
However, the disparity correction with this constant setting ignores the different requirements of offset range for different objects across the whole image. As a matter of fact, the matching errors in nearby slanted surfaces or object boundaries require a large offset for  disparity correction, while the distant discontinuous regions require a much smaller offset for fine adjustment. Moreover, the thin objects like trees and poles even require totally different disparities with the coarse results, because these structures are prone to be damaged in low resolution processing due to the  downsampling and upsampling operations. Accordingly, we propose an efficient dynamic offsets prediction (DOP) module to address the above problems.

The basic idea behind DOP is to dynamically predict the offsets for each pixel according to the guidance of the coarse disparity results and image information. Specifically, we define the offset prediction as a function of the coarse disparity map and the original left image,
\begin{equation}
	\begin{aligned}
		k_{\mathbf{p}}^n=f_{DOP}(\mathbf{I_p}, d_{\mathbf{p}}).
	\end{aligned}
	\label{e2}
\end{equation}
Where $k_{\mathbf{p}}^n$ is the $n$th predicted offset ($k_{\mathbf{p}}^1=0$) and $\mathbf{I_p}$ is the RGB information of pixel $\mathbf{p}$. To obtain the optimal offsets, we propose to approximate the function with a CNN module. Concretely, the proposed DOP firstly resizes the coarse disparity map and the left image to the same dimension. Then, the concatenated disparity map and color image pass through a 2D CNN that outputs a $C_{DOP}$-dimensional representation. Four 2D residual blocks with stride of 1 follow to learn the final predictions. The output of DOP module is a $N$ dimensional offset tensor that is then added to the coarse disparities to generate the adaptive disparity candidates for fine disparity estimation. We note that the  hyper-parameters $N$ and $C_{DOP}$ determine the performance and computation complexity of DOP module. 
\subsection{Disparity-independent convolution}
For the cost aggregation step in previous works,  a series of traditional 3D CNNs with $3\times3\times3$ kernel are exploited to regularize the full-range cost volume and the compact cost volume, where the parameters of the convolutional kernel are fully shared among all the disparity candidates in each stage. However, we claim that this weight-shared scheme in disparity dimension is not the optimal operation for the compact cost volume regularization due to its different local statistical characteristics compared to that of the full-range cost volume. In the coarse disparity estimation stage, all the disparity candidates in full search range have the same probability of being the optimal result. Nevertheless, the $N$ disparity candidates in the fine disparity estimation stage actually have distinct probabilities since the coarse results in the previous stage already provide a priori information for the disparity correction. 
The traditional 3D CNNs which share weights along disparity dimension cannot  learn distinct representations efficiently among these $N$ candidates. 

Based on the above observation, we propose a disparity-independent convolution (DIC) to address these different characteristics for the compact cost volume. 
Different from the traditional cost aggregation operation, the kernel size of DIC is set as $3\times3\times N$ and the parameters are learned in a weight-unshared manner for the $N$ candidates. It should be noted that the $3\times3\times N$ kernel is used to combine all the information of the $N$ candidates because of their intrinsic relevance, while the weight-unshared scheme is used to learn their independent representations among each other. Furthermore, we find that the proposed DIC can be effectively implemented as 2D CNN through reshaping the $N$ candidates into the channel dimension, thus it also regularizes the compact cost volume in spatial and disparity dimension simultaneously. Therefore, the proposed module largely improves the disparity accuracy as well as the computation speed.
\subsection{Disparity regression and loss function}
For coarse disparity regression in stage 1, we use the soft argmin scheme proposed in~\cite{kendall2017end}, which is fully differentiable and enables sub-pixel estimation. The coarse disparity $\hat{d}_1$ is the sum of all the disparity candidates weighted by their probabilities
\begin{equation}
	\begin{aligned}
		\hat{d}_1=\sum_{d=0}^{D_{max} -1} d \times \sigma\left(-c_{d}\right).
	\end{aligned}
	\label{e3}
\end{equation}
Where the probability of each disparity $d$ is calculated from the aggregated cost $c_d$ through softmax operation. For stage 2, the disparity $\hat{d}_2$ is calculated in a similar way, except that the candidate disparity is $\hat{d}_1+k^n$
\begin{equation}
	\begin{aligned}
		\hat{d}_2=\sum_{n=1}^{N} (\hat{d}_1+k^n) \times \sigma\left(-c_{\hat{d}_1+k^n}\right).
	\end{aligned}
	\label{e4}
\end{equation}

For loss function, we adopt smooth L1 to supervise the overall network because of its robustness and low sensitivity to outliers. Since each stage has two outputs as mentioned above, the loss is defined as
\begin{equation}
	\begin{aligned}
		Loss=\sum_{s=1}^{2} \sum_{i=1}^{2}  \left(\frac{\lambda_{si}}{P} \sum_{j=1}^{P} smooth_{L_{1}}\left(d_{j}^{*}-\hat{d}_{si j}\right)\right).
	\end{aligned}
	\label{f5}
\end{equation}
Where $\lambda_{si}$ denotes the weight value for $i$th output in stage $s$. $P$ is the number of valid
pixels and $d_{j}^{*}$ is the ground-truth disparity. The smooth L1 is calculated as

\begin{equation}
	\begin{aligned}
		smooth_{L_{1}}(x)=\left\{\begin{array}{ll}
			0.5 x^{2}, & \text { if }|x|<1 \\
			|x|-0.5, & \text { otherwise }.
		\end{array}\right.
	\end{aligned}
	\label{f6}
\end{equation}

\section{Experimental Results}
\subsection{Datasets and evaluation metrics}
Our experiments are conducted on four datasets: Scene Flow~\cite{mayer2016large}, KITTI 2015~\cite{menze2015object}, KITTI 2012~\cite{geiger2012are} and ETH3D~\cite{schops2017multi}. Scene Flow is a synthetic dataset including 39824 images with a resolution of 960$\times$540, in which 35454 samples are used for training and others for testing. Each sample provides a dense disparity ground-truth. KITTI 2012 and KITTI 2015 are two datasets of real-world driving scenarios with a resolution of 1242$\times$375. KITTI 2012 includes 194 training and 195 testing image pairs, while KITTI 2015 contains 200 training and 200 testing image pairs.  Ground-truth disparity labels are sparsely provided by LIDAR. ETH3D offers 20 test and 27 training samples with sparsely labeled ground-truths, which are recorded in a variety of indoor and outdoor scenes.

For Scene Flow datasets, we use end-point error (EPE) as the evaluation metric, which is defined as mean average disparity error in pixels. For KITTI datasets, we follow the evaluation metrics presented in the official website. The results of EPE and D1 in non-occlusion (D1-noc) and all areas  (D1-all)  are reported for KITTI 2015, where D1 is the predicted error which is greater than three pixels and 5\% of the ground-truth label. The results of 3px in non-occlusion (3px-noc) and all areas (3px-all)  are reported for KITTI 2012, where 3px is the predicted error which differs from the label by a threshold of three pixels. For ETH3D datasets, we provide the results of EPE, 2px and A95, where 2px is the fraction of pixels with errors larger than 2 disparities and A95 is the highest disparity error within the 95\% of best pixels.

\subsection{Implementation details}
Our model is implemented in PyTorch framework~\cite{paszke2017automatic} and trained in an end-to-end manner. Before training the proposed ADCPNet,  all the input stereo image pairs are preprocessed using color normalization and randomly cropped to a dimension of 512$\times$256. Our network is firstly trained on Scene Flow for 40 epochs with a constant learning rate of 0.0005. Then, the pre-trained model is finetuned for another 800 epochs on KITTI 2012, KITTI 2015 or ETH3D datasets. The learning rate of the fine-tuning starts from 0.0005 and halves every 200 epochs. Adam optimizer with $\beta_1=0.9$, $\beta_2=0.999$ is used for parameter optimization. The weight values of four outputs are set as $\lambda_{11}=0.25$, $\lambda_{12}=0.5$, $\lambda_{21}=0.5$, $\lambda_{22}=1.0$. The proposed network is trained from scratch with a batch size of 6 on a single NVIDIA RTX 2080Ti GPU. The maximum disparity is set as 192. In addition, KITTI 2015 is split into a ratio of 1:4 for validation and training, and the inference procedure is carried out on both 2080Ti and TX2 with full resolution.

\subsection{Ablation studies}

\subsubsection{Performance evaluation with different settings}
\begin{table}[t]
	\centering
	\begin{threeparttable}
		\resizebox{0.9\columnwidth}{!}{
			\smallskip
			\begin{tabular}{l c c c c }
				\toprule
				\multirow{2}*{Model} & \multirow{2}*{$N$} & Scene Flow & KITTI 2015 & Runtime \\
				&  & EPE(px) & D1-all(\%) & (ms)\\
				\midrule
				\multirow{4}*{Proposed-S}& 3 & 2.890 & 4.40 & 5.5 \\ 
				& 5 & 2.735 & 4.16 & 5.7\\ 
				& 7 & 2.665 & 3.59 & 5.9\\ 
				& 9 & 2.535 & 3.47 & 6.2\\ \midrule
				\multirow{4}*{Proposed-M}&3 & 2.002 & 3.19 & 6.1 \\ 
				& 5 & 1.856 & 2.98 & 6.6 \\ 
				& 7 & 1.789 & 2.90 & 7.0 \\ 
				& 9 & 1.762 & 2.68 & 8.0 \\ \midrule
				\multirow{4}*{Proposed-L}& 3 & 1.570 & 2.47 & 14.7 \\ 
				& 5 & 1.482 & 2.36 & 16.6 \\ 
				& 7 & 1.440 & 2.25 & 18.8 \\ 
				& 9 & 1.429 & 2.12 & 21.4 \\ 
				\bottomrule
		\end{tabular}}
		\smallskip
	\end{threeparttable}
	\caption{Ablation studies of the proposed network with different settings on Scene Flow test set and KITTI 2015 validation set. Runtimes are measured on 2080Ti.}
	\label{t1}
\end{table}

\begin{table*}[t!]
	\footnotesize
	\centering
	\scriptsize
	\begin{threeparttable}
		\resizebox{1.85\columnwidth}{!}{
			\smallskip\begin{tabular}{c c c c c c c c}
				\toprule
				\multirow{2}*{Std. 3D CNN} &
				\multirow{2}*{Constant offsets} & \multirow{2}*{DOP} & \multirow{2}*{DIC} & Scene Flow & \multicolumn{2}{c}{KITTI 2015} & Runtime \\
				& & & & EPE(px) & EPE (px) & D1-all (\%) & (ms)\\
				\midrule
				\checkmark & \checkmark & & & 2.399 & 0.866 & 3.63 & 9.2\\
				\checkmark & & \checkmark & & 2.000 & 0.831 & 3.30 & 9.7\\
				& \checkmark & & \checkmark & 2.100 & 0.825 & 3.20 & 6.6\\
				& & \checkmark & \checkmark & 1.789 & 0.789 & 2.90 & 7.0\\
				\bottomrule
		\end{tabular}}
		\smallskip
	\end{threeparttable}
	\caption{Effectiveness evaluations of the proposed modules on Scene Flow test set and KITTI 2015 validation set. The configuration of the first line refers to the baseline model. Runtimes are measured on 2080Ti. }
	\label{t2}
\end{table*}
Since the feature sizes and the number of disparity candidates largely impact the computation
accuracy and speed of the proposed stereo network, we firstly report the experimental results of ADCPNet with different parameter settings on multiple datasets. The proposed network is evaluated at three model scales, namely proposed-S, proposed-M and proposed-L. We set $C$, $C_{3d}$ and $C_{DOP}$ as 4, 8 and 16 respectively for proposed-M model, and multiply them by factors of 0.5 and 2  when scaling to proposed-S and proposed-L models. The number of predicted offsets $N$ is set as 3, 5, 7, 9 for an adequate comparison. As shown in Table \ref{t1}, the computation accuracy gradually improves with the increase of the network size, while the computation speed slows down  accordingly. Similarly, larger $N$ can obtain
better disparity accuracy at the penalty of efficiency degradation. Therefore, the proposed network can be customized to various scales according to different real-time requirements.

\subsubsection{Effectiveness evaluation of the proposed modules}
\begin{figure}[b]
	\centering
	\includegraphics[width=0.90\columnwidth]{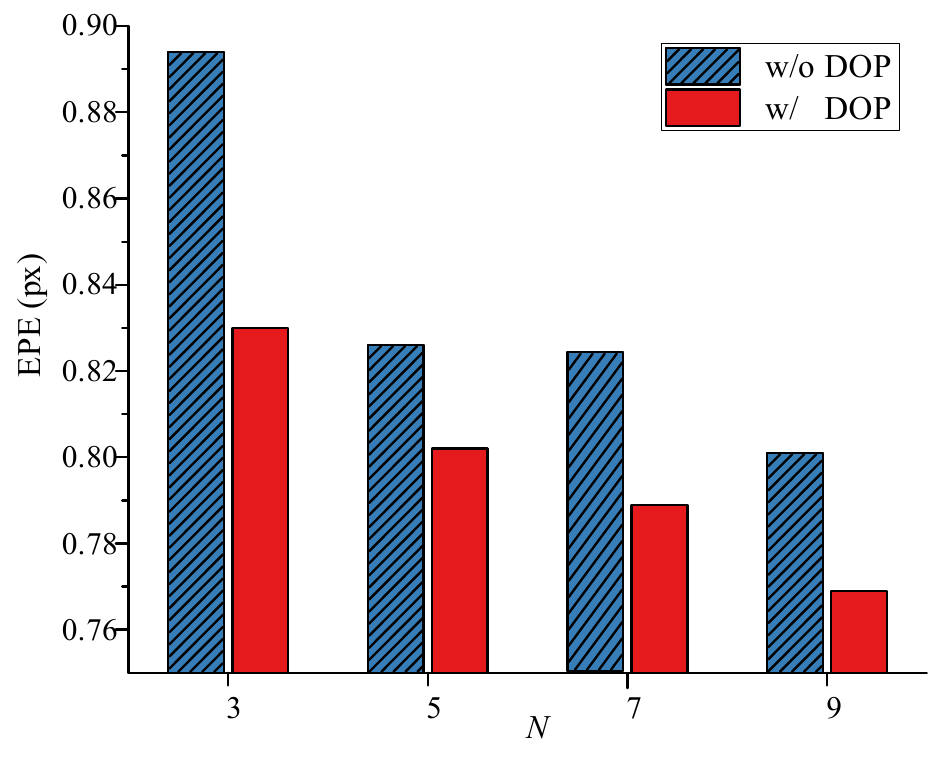} 
	\caption{EPE comparison of KITTI 2015 validation set between the baseline+DIC model (i.e. without DOP) and proposed model (i.e. with DOP) on different $N$.}
	\label{fig6}
\end{figure}

\begin{figure*}[!h]
	\centering
	\includegraphics[width=1.9\columnwidth]{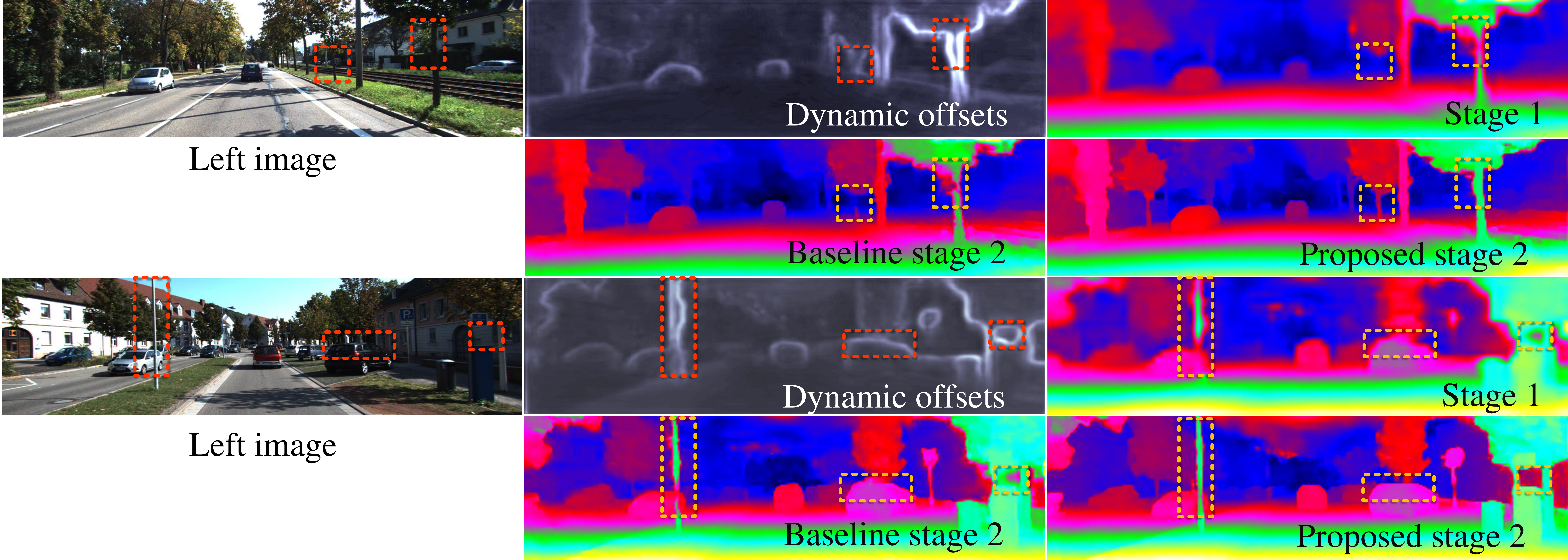} 
	\caption{Subjective comparison of two  samples from KITTI 2015 dataset. Our network enables better disparity estimations for nearby, distant objects and thin structures.}
	\label{fig8}
\end{figure*}
To evaluate the effectiveness of the proposed modules, we firstly build a baseline model which is also a two-stage coarse-to-fine architecture, but applies the constant offsets and standard 3D CNN for compact cost volume regularization. Then, DOP and DIC are gradually deployed into the baseline model for comparison. In this experiment, the hyper-parameters are set as the same as the proposed-M model (here $N$ is set as 7). As shown in Table \ref{t2}, the disparity accuracy can be largely improved by using DOP or DIC module individually, and DIC also reduces the inference time by 28\% because of the simple 2D CNN reshaping operation. By integrating these two modules, the proposed network increases the accuracy and speed by 20\% and 24\% respectively, compared with the baseline model. Figure \ref{fig6} further illustrates the EPE comparison results of KITTI 2015 between the baseline+DIC model (i.e. without DOP) and proposed model (i.e. with DOP) on different number of disparity candidates $N$. We can see that DOP enables obvious accuracy improvement in terms of each setting, and it can maintain a high accuracy even with only 3  candidates. 

The subjective comparison of two samples from KITTI 2015 is illustrated in Figure \ref{fig8}. As it can be observed, the predicted dynamic offsets are totally adaptive to various regions. Therefore, the coarse matching errors can be corrected precisely especially in the discontinuous areas, and the accurate disparity of the thin structures even can be recovered by the proposed method. See more subjective comparisons of KITTI and Scene Flow datasets in the supplementary materials.
\subsubsection{Efficiency evaluation of the proposed network}
\begin{table}[b]
	\centering
	\begin{threeparttable}
		\resizebox{1.0\columnwidth}{!}{
			\begin{tabular}{l c c c }
				\toprule
				\multirow{2}*{Model} & Scene Flow & KITTI 2015 & Runtime\\
				& EPE (px) & D1-all (\%) &(ms)\\
				\midrule
				Baseline & 2.064 & 2.74 & 19.9\\ 
				Baseline+1/8 stage & 1.828 & 2.36 & 23.8\\ 
				Proposed-L & 1.482 & 2.36 & 16.6\\  
				Proposed-L+1/8 stage & 1.485 & 2.30 & 20.3\\ 
				\bottomrule
		\end{tabular}}
		\smallskip
	\end{threeparttable}
	\caption{Efficiency evaluations of the proposed network with different number of stages on Scene Flow test set and KITTI 2015 validation set. Here $N$ is set as 5 for each model. Runtimes are measured on 2080Ti.}
	\label{t3}
\end{table}
We add an extra 1/8 stage to baseline and proposed-L respectively to demonstrate the efficiency of the proposed network. As shown in Table~\ref{t3}, the three-stage design significantly improves the accuracy of the two-stage baseline model at the cost of efficiency degradation. However, the two-stage proposed-L model already outperforms the  three-stage baseline model by notable margins on all evaluation metrics with less inference time. This is mainly because the proposed DOP predicts dynamic offsets and thus generates adaptive disparity candidates for each pixel. 
Moreover, an additional processing stage cannot achieve obvious accuracy improvement when applied on the proposed network. As a matter of fact, the evaluation results on Table~\ref{t1} and Table~\ref{t3} show that increasing the candidate number is more efficient than increasing the processing stage. These evaluations further validate our claim that  the coarse matching errors can be corrected efficiently with fewer stages if we can provide  more accurate disparity candidates.

\subsection{Comparison with SOTA networks}

\subsubsection{Performance comparison on TX2}
\begin{figure}[t]
	\centering
	\includegraphics[width=0.9\columnwidth]{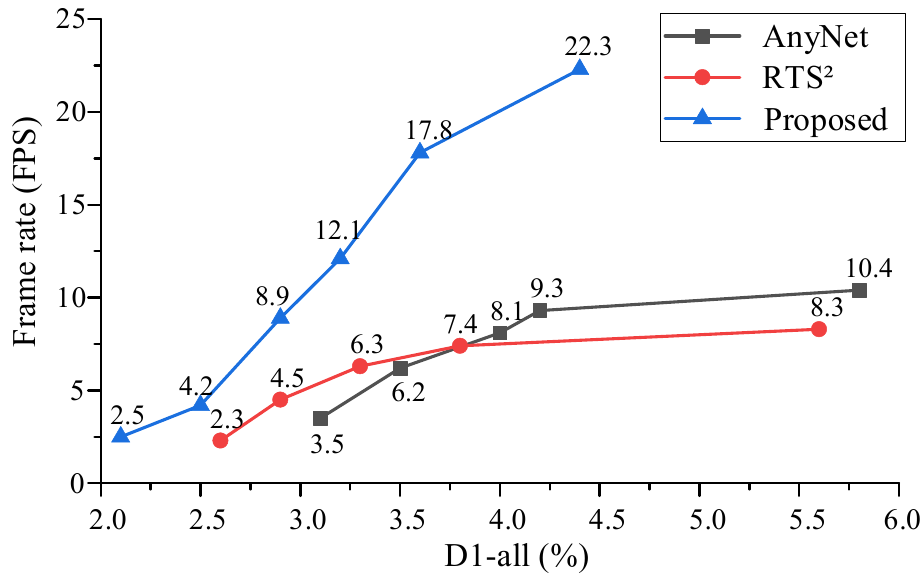} 
	\caption{Accuracy and frame rate comparisons between the proposed design and SOTA networks on TX2.}
	\label{fig5}
\end{figure}

To demonstrate the superiority of the proposed ADCPNet on mobile device, we compare our design with AnyNet~\cite{yan2019anytime} and RTS$^2$~\cite{dovesi2019real} which currently achieve the highest frame rate and accuracy respectively on TX2 platform. The D1-all results on KITTI 2015 validation set and the frames per second (FPS) of these three models are illustrated in Figure \ref{fig5}. It should be noted that RTS$^2$ and AnyNet both exploit elaborate refinement operations to  improve their network outputs, and 
the results of different settings of these two models are directly presented in~\cite{dovesi2019real}. As it can be seen, the proposed network achieves the highest inference frame rate 22 FPS for the final stage disparity estimation with much lower D1-all result. It is $2\times$ faster than AnyNet and $3\times$ faster than RTS$^2$. For the disparity accuracy comparison, the proposed network without any refinement operation improves the previous SOTA RTS$^2$ results  by 19\% because of the effectiveness of the proposed modules. In addition, in the case of similar computation accuracy, our design at least doubles the computation speed on TX2, compared with AnyNet and RTS$^2$.

\subsubsection{Online results and resource usage comparisons}
\begin{table*}[t]
	\scriptsize
	\centering
	\begin{spacing}{1.}
		\begin{threeparttable}
			\resizebox{1.85\columnwidth}{!}{
				\smallskip\begin{tabular}{l c c c c c}
					\toprule
					\multirow{2}*{Method} & \multicolumn{2}{c}{KITTI 2012} & \multicolumn{2}{c}{KITTI 2015} & Runtime\\
					& 3px-noc (\%) & 3px-all (\%) & D1-noc (\%) & D1-all (\%) & (ms)\\
					\midrule
					GCNet~\cite{kendall2017end} & 1.77 & 2.30 & 2.61 & 2.87 & 900 \\
					PSMNet~\cite{chang2018pyramid} & 1.49 & 1.89 & 2.14 &2.32 & 410 \\
					AcfNet~\cite{you2020adaptive} & 1.17& 1.54& 1.72 & 1.89 & 480 \\
					GANet-15~\cite{zhang2019ga} & 1.36 & 1.80 & 1.73 & 1.93 & 360 \\
					CSPN~\cite{cheng2019learning} &1.19 & 1.53& 1.61& 1.74 & 1000 \\
					PDS~\cite{tulyakov2018practical} & 1.92 & 2.53 & 2.36 & 2.58 & 500 \\
					\midrule
					DispNetC~\cite{mayer2016large} & 4.11& 4.65& 4.05& 4.34 & 60 \\
					DeepPruner-Fast~\cite{duggal2019deep} & 2.26* & 2.73* & 2.35& 2.59 & 60 \\
					StereoNet~\cite{khamis2018stereo} & -& -& -& 4.83 & 15 \\
					MADNet~\cite{tonioni2019real} & 5.34*& 6.39*& 4.27& 4.66 & 20 \\
					RTS$^2$~\cite{dovesi2019real} & -& -& 3.22& 3.56 & 20 \\
					AnyNet ($C=1$)~\cite{yan2019anytime} & 5.53* & 6.42* & 6.66* & 7.10* & 8* \\
					AnyNet ($C=32$) & 3.60* & 4.33* & 4.63* & 4.95* & 13* \\
					Proposed-S ($N=3$) & 4.86 & 5.70 & 5.33 & 5.70 & 6  \\
					Proposed-M ($N=7$) & 3.14 & 3.84 & 3.69 & 3.98 & 7  \\
					Proposed-L ($N=9$) & 2.44 & 3.04 & 2.84 & 3.09 & 21 \\
					Proposed-L+refine & 2.20 & 2.66 & 2.52 & 2.71 & 27 \\
					\bottomrule
			\end{tabular}}
		\end{threeparttable}
	\end{spacing}
	\smallskip
	\caption{Results on KITTI online benchmark. * indicates the results which are obtained from the corresponding open source codes, and - indicates that there is no available results or open source codes for evaluation. Runtimes are measured on high-end GPUs.}
	\label{t4}
\end{table*}

We compare the test results of the proposed network with SOTA large-scale and lightweight stereo networks on KITTI 2012 and 2015 online benchmarks. To make a fair comparison, we also employ a popular refinement operation for the proposed design which is widely used in the existing lightweight stereo networks~\cite{khamis2018stereo,duggal2019deep}. As shown in Table \ref{t4},  large-scale models do yield high disparity accuracy, but the inference time is far from achieving real-time performance. In comparison, the lightweight models address the speed limitations with acceptable performance on diverse GPU platforms. Table \ref{t5} further presents the resource usage of various lightweight stereo networks in terms of FLOPs, parameters and memory consumption.  Among these efficient designs, we can observe that the proposed network achieves higher computation speed  on both TX2 and 2080Ti, and the accuracy of our work largely exceeds the previous coarse-to-fine designs~\cite{tonioni2019real,dovesi2019real,yan2019anytime} and is even comparable to SOTA DeepPruner~\cite{duggal2019deep}. Therefore, the proposed ADCPNet is an efficient solution for real-time stereo matching especially for mobile devices.

Finally, we present the online results on ETH3D datasets in Table \ref{t6}. Compared to SOTA lightweight stereo networks~\cite{tonioni2019real,yan2019anytime,duggal2019deep}, the proposed method consistently achieves better accuracy with similar or faster speed, which further demonstrates the high efficiency of the proposed method. The subjective comparisons are included in the supplementary materials.

\begin{table}[t]
	\centering
	\begin{threeparttable}
		\resizebox{1.0\columnwidth}{!}{
			\smallskip\begin{tabular}{l c c c c }
				\toprule
				\multirow{2}*{Model} & FLOPs & Params & Memory & Frame rate \\ 
				& (G) & (M) & (MB)& (FPS) \\ \midrule
				DeepPruner-Fast & 194.24 & 7.47 & 1303 & 0.68 \\ 
				StereoNet & 95.68 & 0.62 & 1195 & 0.70 \\
				RTS$^2$ & 76.93 & 10.35 & - & 2.30 \\ 
				MADNet & 55.66 & 3.82 & 1141 & 3.85 \\  
				AnyNet ($C=32$) & 32.50 & 1.95 & 1363 & 3.50 \\
				AnyNet ($C=1$) & 1.34 & 0.04 & 855 & 10.40 \\
				\midrule
				Proposed-S ($N=3$) & 1.21 & 0.05 & 907 & 22.32 \\ 
				Proposed-M ($N=7$) & 10.24 & 0.43 & 973 & 8.93 \\ 
				Proposed-L ($N=9$) & 77.64 & 3.56 & 1163 & 2.49 \\ 
				Proposed-L+refine & 94.54 & 3.68 & 1169 & 1.89 \\
				\bottomrule
		\end{tabular}}
	\end{threeparttable}
	\smallskip
	\caption{Resource usage and frame rate comparisons among diverse lightweight stereo networks on KITTI 2015 dataset, where the frame rate is reported on TX2.}
	\label{t5}
\end{table}

\begin{table}[t]
	\centering
	\begin{threeparttable}
		\resizebox{0.9\columnwidth}{!}{
			\smallskip\begin{tabular}{l c c c c}
				\toprule
				\multirow{2}{*}{Method} & EPE  & 2px & A95  &Runtime \\ 
				& (px) & (\%) & (px)& (ms)\\
				\midrule
				AnyNet ($C=1$) & 0.68 & 7.15 & 2.38 & 14 \\
				AnyNet ($C=32$) & 0.55 & 4.41 & 2.14 & 37 \\
				MADNet & 0.62 & 4.31 & 1.80 & 63 \\  
				DeepPruner-Fast & 0.46 & 2.71 & 1.68 & 179 \\ 
				\midrule
				Proposed-S ($N=3$) & 0.62 & 5.55 & 2.35 & 8 \\ 
				Proposed-M ($N=7$) & 0.53 & 4.33 & 2.11 & 17 \\ 
				Proposed-L ($N=9$) & 0.50 & 3.64 & 1.82 & 65 \\ 
				Proposed-L+refine & 0.41 & 2.57 & 1.50 & 76 \\
				\bottomrule
		\end{tabular}}
	\end{threeparttable}
	\smallskip
	\caption{Comparisons with lightweight stereo networks on ETH3D online benchmark. The results of existing methods are estimated based on their open source codes. Runtimes are measured on 2080Ti. }
	\label{t6}
\end{table}

\section{Conclusion}
This work concerns the efficient and lightweight stereo network model for accurate and rapid disparity map estimation. In particular, we design a two-stage coarse-to-fine framework and propose a dynamic offset prediction and a disparity-independent convolution to improve the computation accuracy as well as the speed. According to the extensive evaluations on Scene Flow, KITTI and ETH3D datasets and 2080Ti, TX2 platforms, the proposed stereo network enables faster and more accurate disparity estimation especially for mobile devices, compared with the previous SOTA lightweight models. Moreover, the proposed two modules can be taken as plug-and-play components to improve the real-time performance of the existing coarse-to-fine designs.

{\small 
	\bibliographystyle{ieee_fullname}
	\bibliography{1}
}

\end{document}